\documentclass[letterpaper, 10 pt, conference]{ieeeconf}  

\usepackage{graphicx}
\usepackage{amsmath}
\usepackage{amsfonts}

\IEEEoverridecommandlockouts                              

\title{\LARGE \bf
Acoustic Feedback for Closed-Loop Force Control in Robotic Grinding
}

\author{
Zongyuan Zhang$^{1, 2, 3, *}$, Christopher Lehnert$^{1, 2, 3}$, Will N. Browne$^{1, 2, 3}$, Jonathan M. Roberts$^{1, 2, 3}$
\thanks{$^{1}$School of Electrical Engineering and Robotics, Queensland University of Technology, 2 George St, Brisbane, 4000, Queensland, Australia}%
\thanks{$^{2}$Australian Cobotics Centre, Queensland University of Technology, 2 George St, Brisbane, 4000, Queensland, Australia}%
\thanks{$^{3}$ Centre for Robotics, Queensland University of Technology, 2 George St, Brisbane, 4000, Queensland, Australia}
\thanks{$^{*}$Corresponding author: {\tt\small z203.zhang@hdr.qut.edu.au}}%
}

\begin{document}
\bstctlcite{bstctl:forced_etal,bstctl:nodash}

\maketitle
\thispagestyle{empty}
\pagestyle{empty}

\begin{abstract}

Acoustic feedback is a critical indicator for assessing the contact condition between the tool and the workpiece when humans perform grinding tasks with rotary tools. In contrast, robotic grinding systems typically rely on force sensing, with acoustic information largely ignored. This reliance on force sensors is costly and difficult to adapt to different grinding tools, whereas audio sensors (microphones) are low-cost and can be mounted on any medium that conducts grinding sound.

This paper introduces a low-cost Acoustic Feedback Robotic Grinding System (AFRG) that captures audio signals with a contact microphone, estimates grinding force from the audio in real time, and enables closed-loop force control of the grinding process. Compared with conventional force-sensing approaches, AFRG achieves a 4-fold improvement in consistency across different grinding disc conditions. AFRG relies solely on a low-cost microphone, which is approximately 200-fold cheaper than conventional force sensors, as the sensing modality, providing an easily deployable, cost-effective robotic grinding solution.

\end{abstract}

\section{INTRODUCTION}
Grinding and other surface treatment processes (e.g., sanding and polishing) are essential in a wide range of manufacturing domains, including vehicle manufacturing, mould fabrication, and artwork processing. These processes achieve high-quality surface finishes and geometric accuracy by consistently removing small amounts of material from the workpiece surface~\cite{Kopac2006-kg}. Compared with sanding or polishing, grinding places a stronger emphasis on achieving precise dimensional accuracy. However, manual grinding is time-consuming, requires skilled operators, and generates fine dust particles that pose significant occupational health risks. In recent years, robotic grinding has attracted increasing attention in modern manufacturing due to its ability to perform tasks automatically, efficiently, and reliably, delivering reliable dimensional accuracy while reducing labour reliance and health risks~\cite{robotic_sanding_overview}.

To ensure high-quality and consistent results, robotic grinding relies heavily on closed-loop force control. In practice, the normal grinding force between the manipulator's end-effector and the workpiece is most commonly used as feedback for this purpose. Measuring this force typically requires installing force sensors between the manipulator and the grinding tool, which introduces installation constraints. To address this, Alatorre et al.~\cite{Alatorre2020-ho} employed a free-field microphone array to predict grinding force in narrow-space tasks. However, our experiments show that this grinding model-based approach is not generalisable to low-cost microphones and cannot robustly estimate grinding force in the presence of ambient noise.

\begin{figure}[tbp] 
    \centering
    \includegraphics[width=\linewidth]{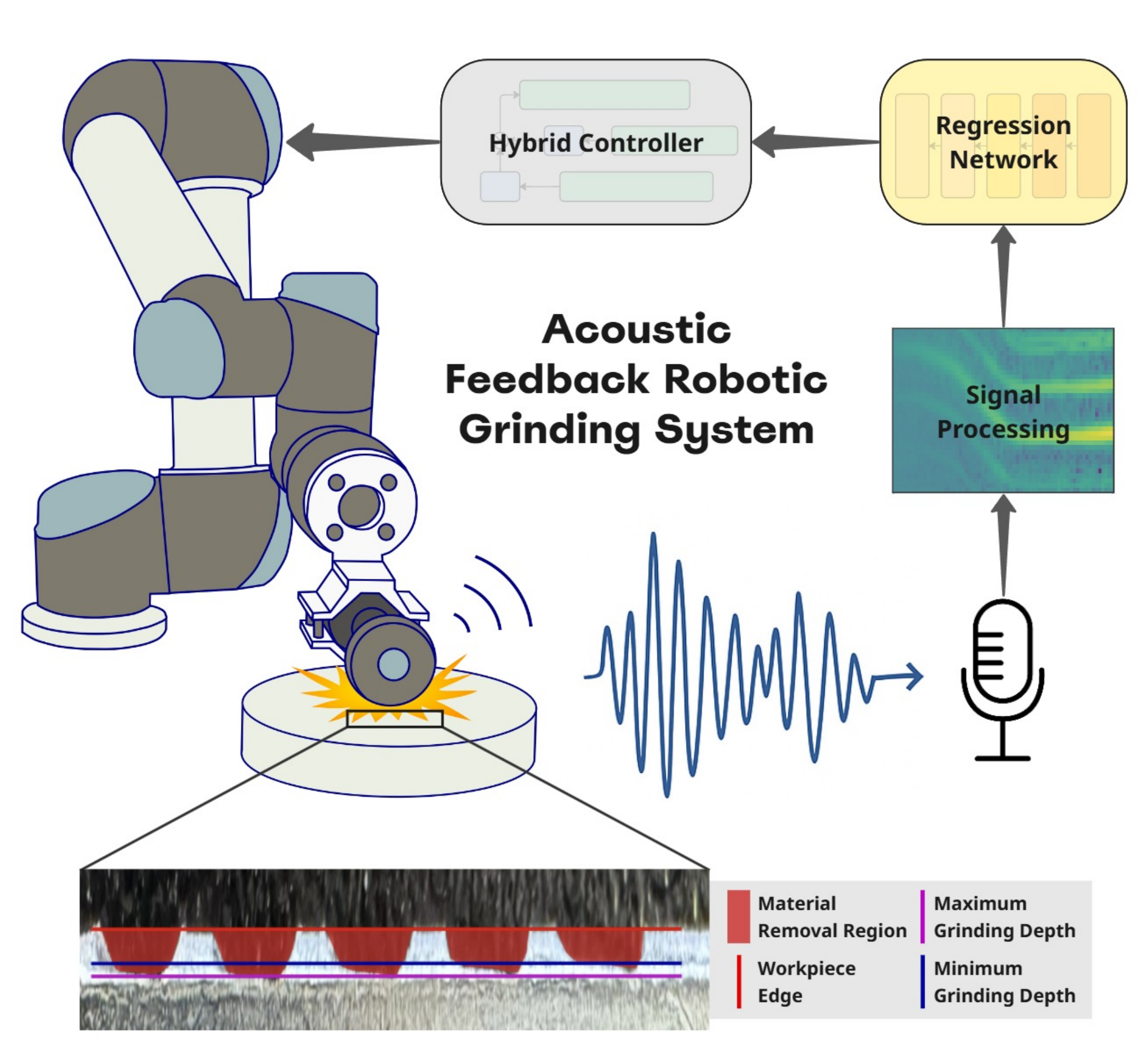}
    \caption{Acoustic Feedback Robotic Grinding System (AFRG) uses acoustic signals for closed-loop force control in robotic grinding. Instead of expensive force sensors, AFRG employs a cost-effective contact microphone to estimate the grinding force. The workflow involves audio recording, signal processing, and regression to estimate the force, which is then used as feedback to regulate the grinding process. An example of the grinding result is shown below the figure, where the uniform grinding depth demonstrates a consistent Material Removal Rate (MRR).}
    \label{fig:overview}
\end{figure}

To overcome the limitations of previous methods, we propose a data-driven Acoustic Feedback Robotic Grinding System (AFRG), as shown in Fig.~\ref{fig:overview}. A Convolutional Neural Network (CNN) is employed to extract acoustic features and estimate grinding force from signals collected by a single contact microphone, which is then used for closed-loop control of the grinding process.

Compared with Force/Torque (F/T) sensors (4,400\,USD) and free-field microphone arrays (3,800\,USD), contact microphones are highly cost-effective (20\,USD), lightweight, and easy to install. The AFRG is also compatible with existing force controllers, allowing direct integration into current robotic platforms. Experimental results show a 4-fold increase in Material Removal Rate (MRR) consistency across varying disc conditions and robust performance under ambient noise.

The key contributions of this work can be summarised as follows:

\begin{enumerate}
    \item \textbf{A data-driven robotic grinding system} that achieves stable and consistent material removal rates, bypassing the need for predefined modelling.

    \item \textbf{A 2D CNN-based acoustic--force estimation method} that performs real-time force estimation from raw audio signals, remaining robust to ambient noise.
    
    \item \textbf{A low-cost sensing approach} that relies solely on a single contact microphone, providing approximately a 200-fold cost reduction.

    \item \textbf{Experimental validation} of this method showing a 4-fold increase in MRR consistency compared to conventional force-sensing approaches.
\end{enumerate}



    
    

\section{BACKGROUND}
Grinding is generally divided into three stages: friction, ploughing, and cutting, with only the cutting stage actually removing material~\cite{Li2017-gq}. In robotic grinding, to enable effective cutting, the normal grinding force must be maintained above a threshold, but not excessively high to avoid workpiece burning. Within a reasonable range of normal force, the force magnitude must also be controlled to indirectly regulate the material removal rate (MRR), thereby ensuring uniform dimensional accuracy.

Early work by Liu et al.~\cite{liu1990robotic} employed PID-based force control with pre-programmed robot trajectories, while Nagata et al.~\cite{Nagata2007-gl} proposed an impedance control~\cite{Hogan1985-rk} based framework for precise surface following on complex furniture surfaces. More recent studies include Maric et al.~\cite{Maric2020-qm}, who designed a framework based on forward dynamics compliant control~\cite{Scherzinger2017-gq} that incorporates human grinding experience. Alt et al.~\cite{Alt2024-eo} integrated computer vision and voice interaction to automatically plan grinding paths, which were executed via hybrid force–position control~\cite{Raibert1981-il}. Despite differences in control strategy, these force-control approaches all rely on accurate force sensors to achieve high grinding quality.

 Accurate force sensors are often expensive, which motivates alternatives such as the virtual sensing approach proposed by Yen et al.~\cite{Yen2019-kw} to reduce costs for collaborative manipulators. Besides sensor cost, force measurements are contaminated by high-frequency vibrations of the grinding tool. To suppress this noise, Maric et al.\cite{Maric2020-qm} applied tailored linear filters, although this also removes high-frequency information. Nogi et al.~\cite{Nogi2022-fi} use high-frequency components of the force data to mitigate various sensor errors, such as offset~\cite{Kim2020-wp} and thermal drift~\cite{Chavez2019-nm}. They perform a Fast Fourier Transformation (FFT) on the force variations and use a Long Short-Term Memory (LSTM) network~\cite{hochreiter1997long} for real-time error mitigation. However, microphones offer easier access to high-frequency signals at a lower cost than force sensors, leading some studies to analyse the grinding process using acoustic measurements.

Bhandari~\cite{Bhandari2021-fk} summarised methods that combine microphones with force sensors to predict post-grinding surface roughness, while Zhang et al.\cite{Zhang2018-cr} and Lee et al.\cite{Lee2020-pn} employed acoustic signals to monitor tool wear of sanding belts and grinding disks, respectively. These studies demonstrate that acoustic signals provide rich information during grinding. Shi et al.\cite{Shi2023-lz} further investigated the use of acoustic signals in human–robot interaction for telerobotic machining tasks, where an acoustic signal was captured and converted into force information to provide haptic feedback to the operator. Kanda et al.\cite{Kanda2024-zq} systematically investigated the relationship between grinding force and the resulting acoustic signals, showing that the fundamental frequency is closely related to grinding force and tool rotation speed, whereas the sound pressure level correlates more strongly with tool wear. 

By exploiting the relationship between grinding force and the fundamental frequency of the acoustic signal, Alatorre et al.~\cite{Alatorre2020-ho} implemented closed-loop force control using acoustic features. Audio captured by an array of free-field microphones was processed via Short-Time Fourier Transform (STFT), and the fundamental frequency was tracked as the dominant spectral peak across frames. The fundamental frequency was mapped through a pre-calibrated equation to the grinding force for feedback regulation. The method relies on the array’s spatial directivity and uniform sound pressure capture for stable extraction of the fundamental frequency, and on the accuracy of the pre-calibrated model of the grinding process for correct force estimation.


\section{PROBLEM FORMULATION}

The objective of the grinding operation is to remove a layer of material from the surface of the workpiece. For common grinding tasks, such as removing surface oxides, it is desirable to remove a uniform amount of material at each location on the workpiece. Achieving this requires closed-loop control to stabilise the MRR across the workpiece. However, since the MRR is difficult to measure in real time, previous work has typically controlled the normal force during grinding. This strategy is supported by the Preston equation~\cite{Preston1927-ls}:
\begin{equation}
\label{eq:MRR}
\dot{V} = \frac{k_p F_n v_r}{A},
\end{equation}
where $\dot{V}$ is the MRR, defined as the volume of material removed per unit time;
$F_n$ is the normal grinding force; 
$v_r$ is the relative velocity between the grinding disc and the workpiece; 
$A$ is the contact area; 
and $k_p$ is the Preston coefficient, which depends on the type of material, the condition of the grinding disc, etc.

\begin{figure}[htbp] 
    \centering
    \includegraphics[width=1\linewidth]{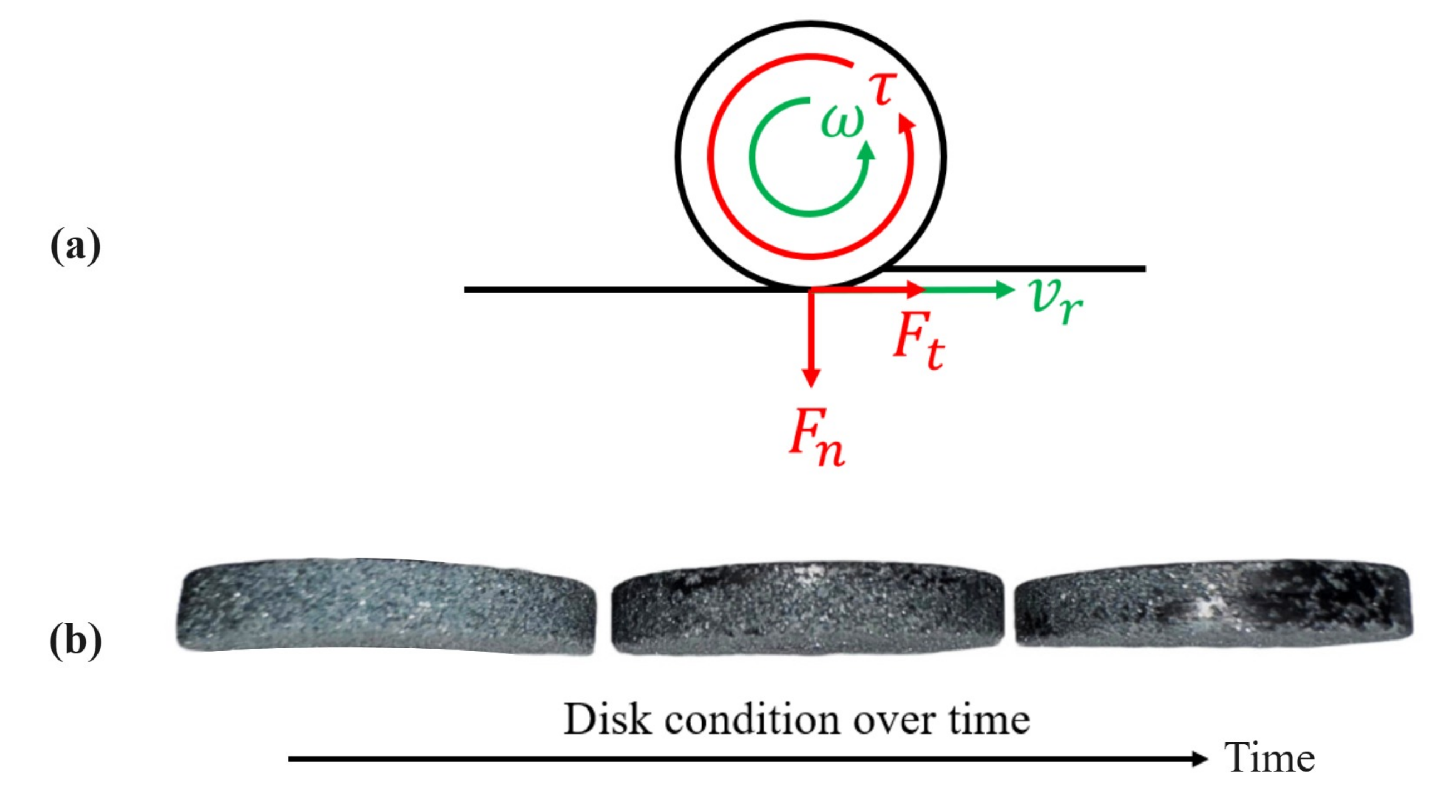}
    \caption{(a) The forces, torque, and velocity of the tool relative to the workpiece surface (side view). (b) Different conditions of the grinding disc on the rotary tool during operation (top view). }
    \label{fig:tool_life}
\end{figure}

Li et al.~\cite{Li2024-tl} modelled the microscopic abrasive grains on a grinding disc and found that, under a fixed disc condition, the ratio between the tangential force $F_t$ and the normal force $F_n$ (as shown in Fig.~\ref{fig:tool_life}(a)) remains approximately constant:

\begin{equation}
\label{eq:engagement}
\mu_\theta = \frac{F_t}{F_n},
\end{equation}
where $\mu_\theta$ is the engagement coefficient between the abrasive tool and the workpiece material, which is an empirical parameter typically determined through experiments, and $\theta$ denotes a specific disc condition.

For a rotary tool equipped with a grinding disc, the force diagram is illustrated in Fig.~\ref{fig:tool_life}(a). The angular velocity of the grinding disc $\omega$ depends on the motor torque $\tau$ through a monotonic DC motor torque-speed function:
\begin{equation}
\label{eq:motor}
    \tau = f_m(\omega).
\end{equation} 
Assuming $f_m(\cdot)$ is invertible, since $\tau = F_t r$ with $r$ being the disc radius and $F_t = \mu_\theta F_n$, $v_r$ can be expressed as a function of $F_n$: 
\begin{equation}
\label{eq:v_r}
v_r = r \omega = r f_m^{-1}(\mu_\theta F_n r).
\end{equation}
By substituting Eq.~(\ref{eq:v_r}) into Eq.~(\ref{eq:MRR}), the MRR becomes: 
\begin{equation}
\dot{V} = \frac{k_p F_n \, r \, f_m^{-1}(\mu_\theta F_n r)}{A},
\label{eq:mrr_fn}
\end{equation}
showing that, under the conditions of constant $A$, $r$, $\mu_\theta$, and $k_p$, the MRR is a function of $F_n$. 
Therefore, closed-loop control of $F_n$ is sufficient to stabilise the MRR.

The proposed system AFRG uses an acoustic signal $s$ to control the grinding process. The goal is to estimate $F_n$ from the measured $s$, so that the existing force-feedback controller can regulate MRR using the estimated normal force $\hat{F}_n$. 

Kanda et al.~\cite{Kanda2024-zq} showed that the $s$ generated during grinding is closely related to $\omega$, as the rotation of the tool is the dominant vibration source in the system. Hence, if $\omega$ can be inferred from $s$, it becomes possible to estimate $F_n$ by linking $\omega$ to $F_n$. Based on Eqs.~(\ref{eq:engagement}) and~(\ref{eq:motor}), the relationship between $F_n$ and $\omega$ can be expressed as
\begin{equation}
F_n = \frac{F_t}{\mu_\theta} = \frac{\tau}{r \mu_\theta} = \frac{f_m(\omega)}{r \mu_\theta}.
\label{eq:w-fn}
\end{equation} 
Using this relationship, we define a composite estimator $x(\cdot)$, which implicitly includes the mapping $g(\cdot)$, as
\begin{equation}
\label{eq:x()}
x(s) := \hat{F_n} =  \frac{f_m\big(g(s)\big)}{r \mu_\theta} \,,
\end{equation}
where the estimator first infers the estimation of angular velocity $\hat{\omega} = g(s)$ from $s$, and then maps it to the estimated normal grinding force $\hat{F}_n$ for closed-loop control.

\section{PROPOSED FRAMEWORK}
\begin{figure}[htbp] 
    \centering
    \includegraphics[width=\linewidth]{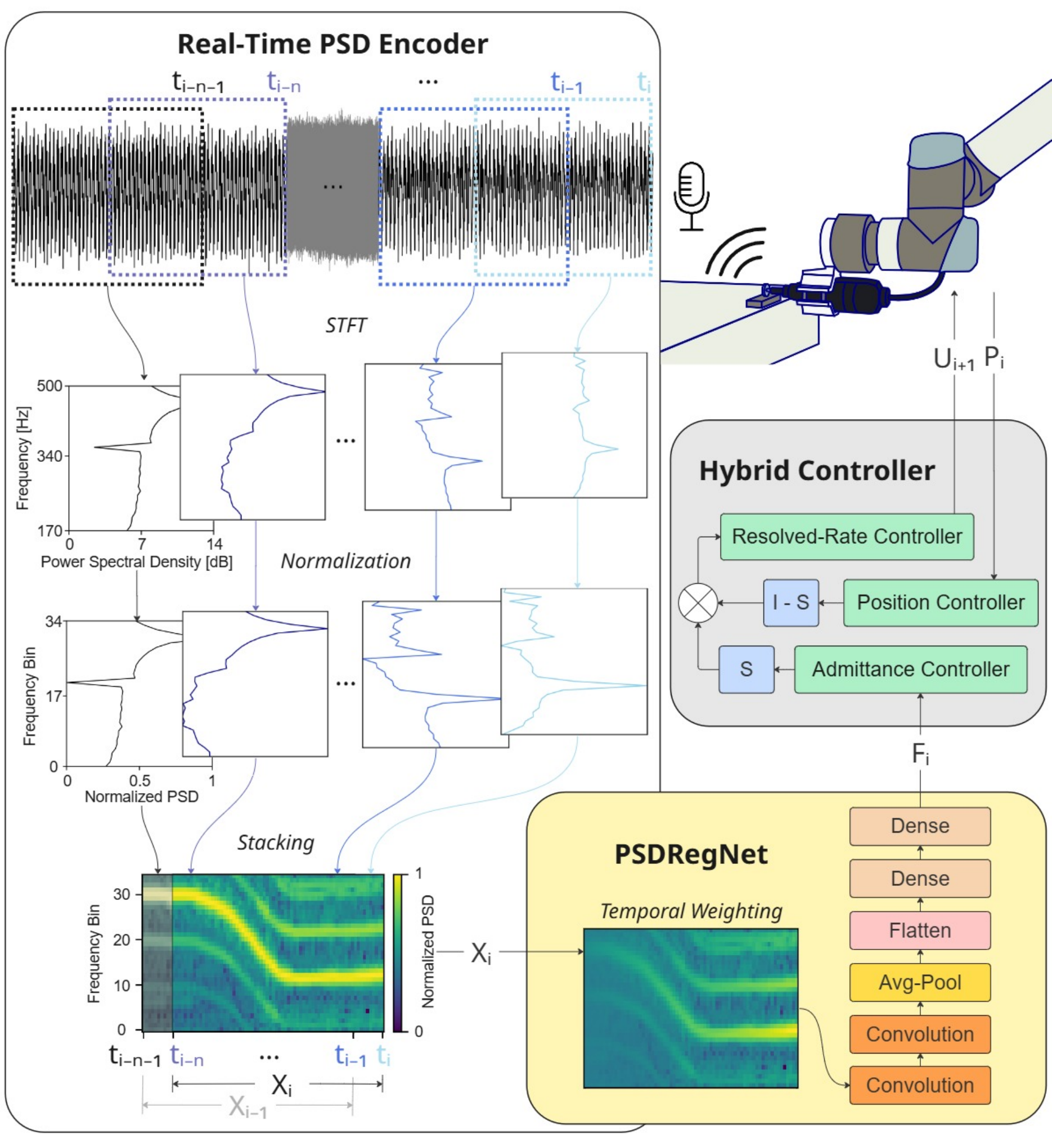}
    \caption{Schematic of the AFRG. It consists of three components: Real-Time PSD Encoder for data processing, PSDRegNet for force estimation, and the Force-Position Hybrid Controller for closed-loop manipulator control. Here, $t_i$ denotes a given time step, $\mathbf{X}_i$ is the stacked PSD encoder output over the $n$-frame receptive field, $F_i$ represents the estimated normal force, $P_i$ is the end-effector position, and $U_i$ is the joint velocity commands.}
    \label{fig:flowchart}
\end{figure}

Based on the problem formulation in the previous section, an estimator $x(\cdot)$ is required to estimate the grinding force from raw acoustic signals. By combining $x(\cdot)$ with a feedback controller, we obtain AFRG, as illustrated in Fig.~\ref{fig:flowchart}. AFRG consists of three main components: a novel real-time Power Spectral Density (PSD) encoder, a tailored PSD Regression Network (PSDRegNet), and an adapted force–position hybrid controller. Audio data are first recorded and pre-processed by the PSD encoder, then passed through PSDRegNet to estimate $\hat{F}_n$, where the PSD encoder and PSDRegNet together constitute $x(\cdot)$. The hybrid controller subsequently uses $\hat{F}_n$ to generate low-level control commands for the manipulator. All components operate at a fixed control frequency $f_c$, producing the next-step joint velocity commands $U_{i+1}$ at each iteration, thereby enabling closed-loop control of the manipulator during grinding.

\subsection{Real-Time PSD Encoder}
In this module, the raw acoustic signal is encoded into a two-dimensional array to facilitate $\omega$ extraction by PSDRegNet.

A common approach to obtain $\omega$ is to transform the acoustic signal from the time domain to the frequency domain using STFT and extract its fundamental frequency \cite{Alatorre2020-ho}. The basic operation of the PSD encoder consists of continuously applying the STFT to newly acquired data and stacking the resulting spectra, thereby providing a continuously updated two-dimensional array in real time.

At each control interval of $1/f_c$, a segment of raw acoustic data of length $t_{ws}$ preceding the current time is extracted as the input for the current time step. Here, $t_{ws}$ denotes the STFT window size, which determines the resulting frequency resolution $\Delta f = 1/t_{ws}$. Short windows lead to insufficient frequency resolution, while long windows may cause loss of short-time features and increase the system latency.

The extracted segment is multiplied by a Hann window\footnote{A tapering window commonly used in spectral analysis to reduce spectral leakage.} and then transformed into the frequency domain via the FFT. The resulting frequency-domain data are converted into a PSD representation, which has been shown to provide higher robustness for fundamental frequency extraction compared to the amplitude spectrum~\cite{Nogi2022-jn}. The PSD is subsequently band-pass filtered along the frequency axis according to the measurable minimum and maximum frequencies of the grinding tool, and normalised along the power axis using min–max normalisation at each time step to mitigate amplitude variations and temporal inconsistencies caused by an uncalibrated microphone and the Analog to Digital Converter (ADC) system.

To form the input for PSDRegNet at a given time step $t_i$, the PSDs from $t_{i-n}$ to $t_i$ are stacked along a time axis into a two-dimensional array $\mathbf{X}_i \in \mathbb{R}^{F \times n}$, where $F$ denotes the number of frequency bins and $n$ is the number of stacked frames corresponding to the network's receptive field in time. 

\subsection{PSDRegNet}
The architecture of PSDRegNet is summarised in Table~\ref{tab:network}, consisting of two convolutional layers followed by two fully connected (FC) layers, with a dropout layer (p=0.3) applied after the first FC layer to prevent overfitting.

\begin{table}[htbp]
\centering
\caption{Network architecture for PSDRegNet}
\begin{tabular}{lcc}
\hline
\textbf{Layer} & \textbf{Output Size} & \textbf{Details} \\
\hline
Input PSD array & $1 \times 35 \times 40$ & -- \\
Conv2D & $16 \times 33 \times 38$ & kernel=5, padding=1 \\
BatchNorm2D & $16 \times 33 \times 38$ & -- \\
ReLU & $16 \times 33 \times 38$ & -- \\
Conv2D & $32 \times 31 \times 36$ & kernel=5, padding=1 \\
BatchNorm2D & $32 \times 31 \times 36$ & -- \\
ReLU & $32 \times 31 \times 36$ & -- \\
AdaptiveAvgPool2D & $32 \times 4 \times 4$ & -- \\
Flatten & $512$ & -- \\
FC & $64$ & Dropout(0.3) \\
ReLU & $64$ & --\\
FC & $1$ & -- \\
\hline
\end{tabular}
\label{tab:network}
\end{table}

PSDRegNet maps the two-dimensional PSD array to the grinding normal force. Conceptually, as described in Eq.~(\ref{eq:x()}), the estimator first infers $\hat{\omega}$ from $s$ and then maps it to $\hat{F}_n$. In practice, we implement this by a 2D CNN that learns $\omega$-related features from the PSD array, followed by the fully connected layers that regress the final output $\hat{F}_n$.

Since the angular speed of the rotary tool changes continuously, the fundamental frequency in the PSD array representing $\omega$ also varies continuously over time. This fundamental frequency typically exhibits the strongest intensity, as illustrated by the bright horizontal streak in Fig.~\ref{fig:dataset}. This property helps the CNN suppress irregular, non-continuous environmental noise or resonance frequencies, which usually appear as weaker (darker) streaks in Fig.~\ref{fig:dataset}.  

Furthermore, since the regression target is the force value at $t_i$, the PSDs closer to $t_i$ carry more informative features. To emphasise this, we apply a learnable temporal mask $\mathbf{M} \in \mathbb{R}^{n \times n}$ to weight the PSD input:

\begin{equation}
\tilde{\mathbf{X}}_i = \mathbf{X}_i \mathbf{M},
\end{equation}
where $\tilde{\mathbf{X}}_i \in \mathbb{R}^{F \times n}$ is the temporally weighted PSD array. $\mathbf{M}$ is a diagonal matrix with learnable entries $m_1, \dots, m_n$ along the diagonal, which are linearly initialised from 0 to 1 and updated during training.

\subsection{Force–Position Hybrid Controller}
The feedback controller uses the estimated normal force $F_i$ at time $t_i$ (or the measured force signal when an F/T sensor is used) with the end-effector position $P_i$ as feedback signals to generate the joint velocity commands $U_{i+1}$ for the next time step $t_{i+1}$. This closed-loop configuration allows the manipulator to apply the required force to the workpiece while tracking the desired trajectory.


As shown in Fig.~\ref{fig:flowchart}, the controller architecture follows a hybrid force–position control scheme~\cite{Raibert1981-il}. An admittance controller regulates the force along the surface normal axis. We define the corresponding projection matrix as $\mathbf{S} \in \mathbb{R}^{3 \times 3}$. In contrast, a PID controller regulates the end-effector position along the remaining directions, which we denote as the $\mathbf{I}-\mathbf{S}$ directions, where $\mathbf{I}$ is the identity matrix. A resolved-rate controller is used to generate $U_{i+1}$.


\section{EXPERIMENTAL SETUP}
\begin{figure}[htbp] 
    \centering
    \includegraphics[width=0.8\linewidth]{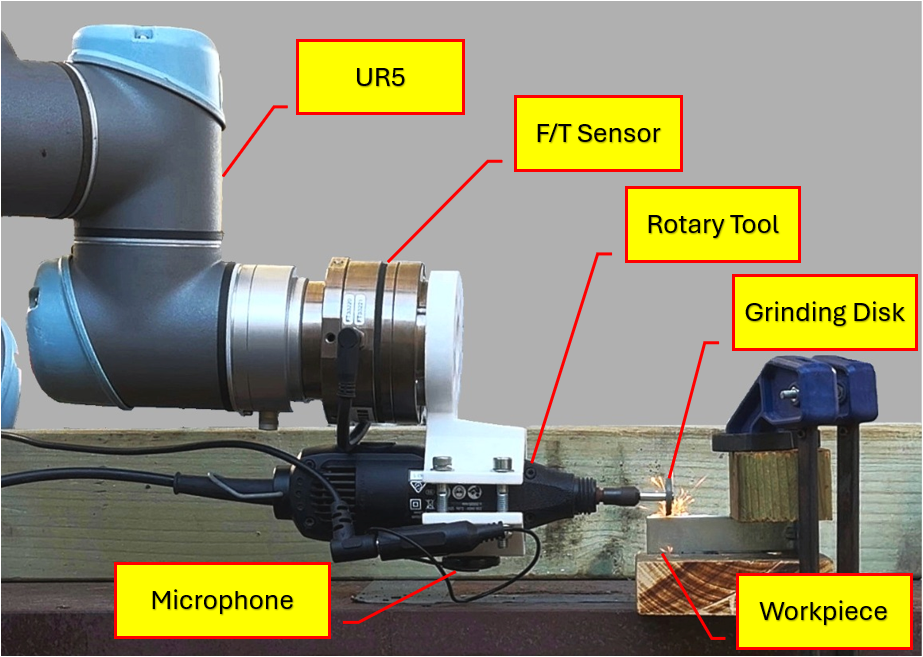}
    \caption{Experimental setup consisting of a 6-DOF UR5 robotic arm, an F/T sensor, a rotary grinding tool with an aluminium oxide disc, a contact microphone, and rectangular workpiece bars made of aluminium alloy or stainless steel. This setup allows grinding to be performed using either F/T sensor-measured forces or acoustically estimated forces, thereby providing a reliable benchmark to evaluate the proposed method.}
    \label{fig:setting}
\end{figure}

The experimental setup used in this study is shown in Fig.~\ref{fig:setting}. A 6-DOF UR5 robotic arm, rigidly fixed to a workbench, manoeuvres a Dremel 200-2/30 rotary tool (125\,W, 35,000\,rpm) with an aluminium oxide grinding disc to perform grinding tasks on the target workpieces. The end-effector is equipped with an Axia80-M20 F/T sensor (ATI Industrial Automation, 200\,N, 8\,Nm range; 0.1\,N, 0.005\,Nm resolution) to provide ground-truth force measurements for validation and training. This configuration enables precise and measurable grinding forces, and these measurements can be directly used for F/T sensor-based closed-loop force control of the grinding task as a baseline.

A low-cost contact microphone, modified from a piezoelectric sensor and mounted on the rotary tool bracket, is connected to an ADC and provides digital audio signals sampled at 96\,kHz for further analysis. Unlike conventional microphones, a contact microphone only picks up vibrations transmitted through solid objects, which helps reduce ambient noise. Although signal quality decreases with increasing distance from the vibration source, the microphone can be flexibly mounted at any location on the robotic arm or workpiece.

The workpieces used in the experiments are rectangular bars made of AA6061 aluminium alloy or SUS304 stainless steel, firmly clamped to the workbench. The majority of experiments and data collection are performed on the aluminium workpieces, which are softer and easier to grind, allowing evaluation of the AFRG under relatively stable disc conditions. The final experiment uses stainless steel, which is harder, causes faster grinding disc wear, and allows assessment of the AFRG under changing disc conditions.

\section{NETWORK TRAINING}
In order to train PSDRegNet to estimate the grinding force, we need a dataset including microphone-recorded acoustic signals and ground truth normal force data measured by the F/T sensor.

\subsection{Data Collection}
The dataset was collected on the experimental system shown in Fig.~\ref{fig:setting} using F/T sensor-based closed-loop force control for the grinding process. Data were collected during fixed-point grinding tasks performed in a real factory environment to naturally introduce ambient noise. The timestamps, force measurements, and audio signals were synchronously recorded.

The dataset, with a total duration of 1400\,s, was collected under two grinding modes: continuous grinding (where the normal force varies continuously from 0\,N to the target value) and intermittent grinding (where the normal force is directly regulated to the target value). Data were collected at five target force levels (2, 3, 4, 5, and 7\,N) for both modes. While ideally the dataset would cover the full range of force values and their corresponding acoustic signals, data collection was limited to forces below 7\,N. This is because forces above 4\,N caused thermal burning of the workpiece and accelerated disc wear, while forces exceeding 7\,N nearly stalled the rotary tool, preventing reliable acoustic feature extraction.

\begin{figure}[htbp] 
    \centering
    \includegraphics[width=1\linewidth]{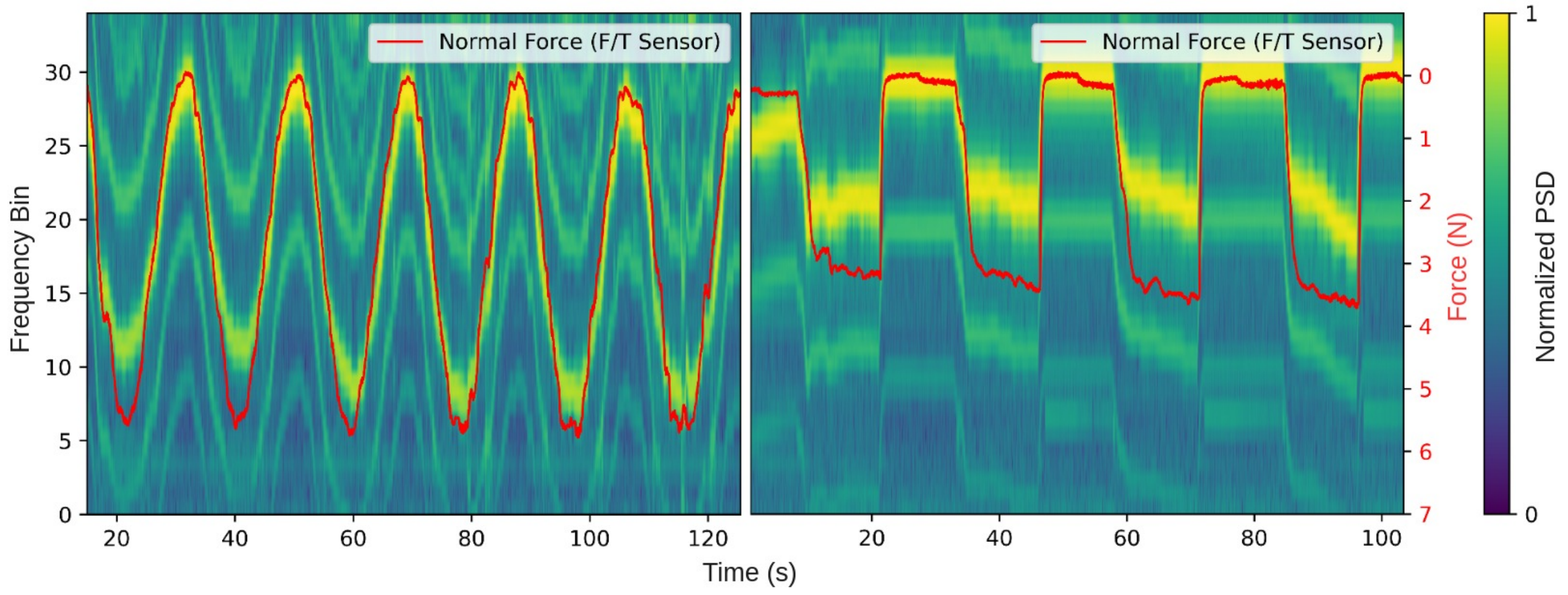}
    \caption{Representative dataset samples are shown for continuous (left) and intermittent (right) grinding. The force axis is inverted to better visualise the correlation between the PSD arrays and the force signal, which are not directly aligned.}
    \label{fig:dataset}
\end{figure}

As shown in Fig.~\ref{fig:dataset}, two representative samples of the collected data in two grinding modes reveal that the relationship between the brightness of the encoded PSD arrays and the measured $F_n$ is a complex, non-linear mapping rather than a direct alignment, highlighting the need for PSDRegNet.

According to Eq.~(\ref{eq:w-fn}), changes in $\mu_\theta$ affect the mapping between $\omega$ and $F_n$. As illustrated in Fig.~\ref{fig:tool_life}(b), the workpiece could experience wear and clogging, which changes $\mu_\theta$. Such tool degradation could be detected by visual inspection and controlled by the timely replacement of the disc.

\subsection{Training Procedure}
The input tensors are constructed by using the PSD encoder to generate a sequence of PSD arrays from raw audio data spanning $t_{i-n}$ to $t_i$. These PSDs are then weighted by a learnable temporal mask $\mathbf{M}$, which is updated via backpropagation. The target labels are the forces measured by the F/T sensor at time $t_i$. The labels are normalised using standard score normalisation (z-score), while the input data are already normalised during PSD encoding and require no further processing. The dataset is split into training and validation sets, and the validation loss is used to guide the learning rate scheduler (ReduceLROnPlateau) during training. The training settings are summarised in Table~\ref{tab:training_settings}.

\begin{table}[htbp]
\centering
\caption{Training settings for PSDRegNet}
\begin{tabular}{l l}
\hline
\textbf{Setting} & \textbf{Value} \\
\hline
Loss function & MSE loss \\
Optimizer & Adam \\
Weight decay & $1\times10^{-4}$ \\
Train/validation split & 80\% / 20\% \\
Number of epochs & 5 \\
Batch size & 64 \\
Initial learning rate & $1\times10^{-4}$ \\
Scheduler & ReduceLROnPlateau, factor=0.5, patience=5 \\
\hline
\end{tabular}
\label{tab:training_settings}
\end{table}

\section{Experimental Results}
\subsection{Force Estimation Performance}
Before applying the AFRG for a grinding operation, it is necessary to validate the force-estimating performance of the PSD encoder and PSDRegNet. This validation is carried out by performing grinding under the control of the F/T sensor while simultaneously estimating the force using PSDRegNet. The force-estimating performance is then evaluated by comparing the measured force with the estimated force.

Two types of grinding tasks are conducted. The first type consists of multiple short grinding trials (10\,s each) at different fixed points, aimed at evaluating the AFRG's transient response under static contact. The second type is a single long grinding trial (90\,s) along a straight-line trajectory, aimed at evaluating the AFRG's ability to sustain consistent grinding force during long-time dynamic operation. Both types are performed under three target forces (2\,N, 3\,N, and 4\,N). These levels are chosen because forces below 2\,N predominantly cause ploughing rather than cutting, while forces above 4\,N accelerate tool wear.

\begin{figure}[htbp] 
    \centering
    \includegraphics[width=0.95\linewidth]{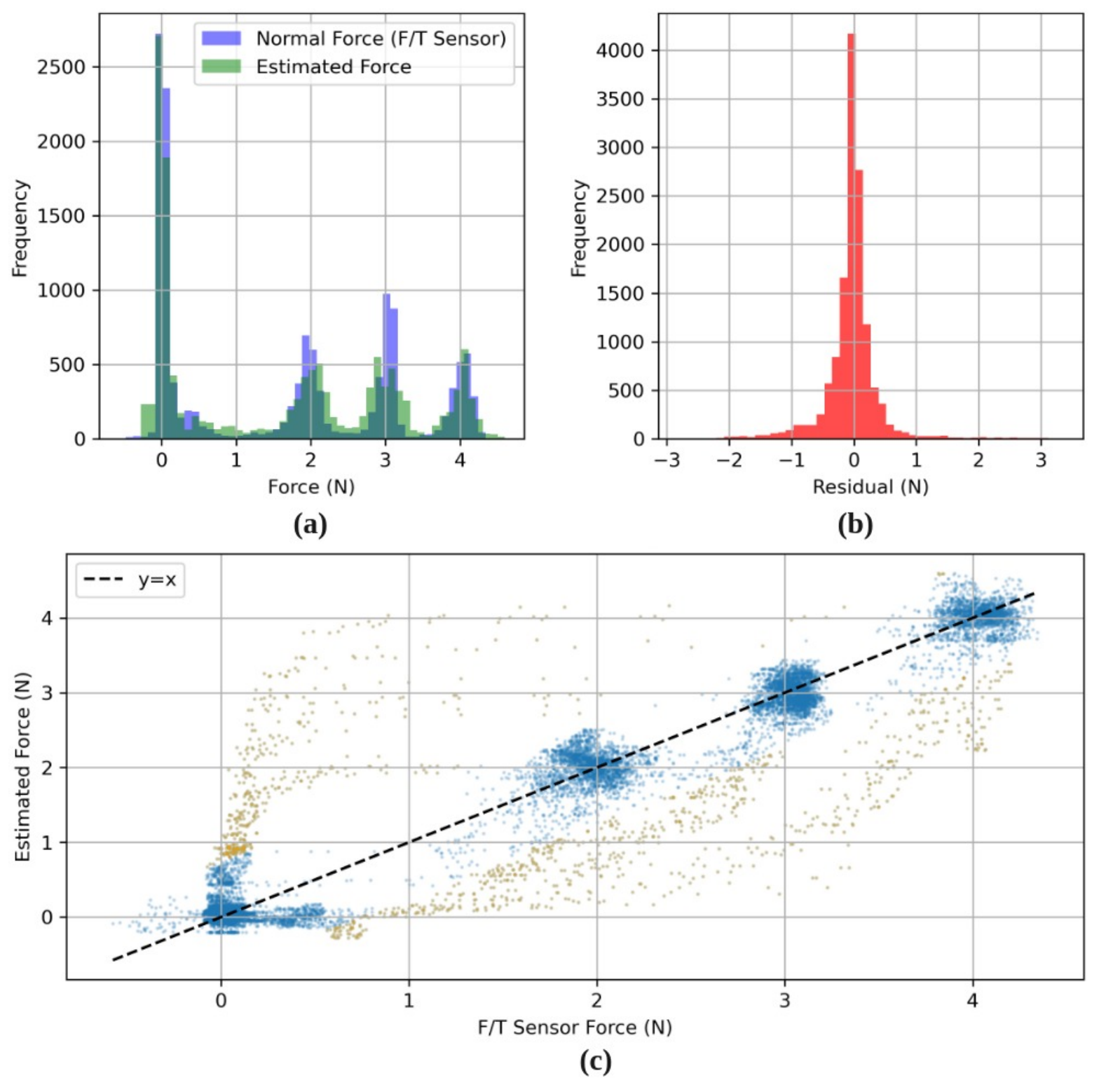}
    \caption{Force estimation results. 
    (a) Distribution of measured and estimated forces, where closer overlap indicates higher accuracy. 
    (b) Residual distribution, which ideally follows a normal distribution. 
    (c) Scatter plot of estimated versus measured forces, where clustering along the $x=y$ line indicates better accuracy. Outliers are highlighted in orange.}
    \label{fig:force_estimating}
\end{figure}

For this experiment, 13,813 pairs of measured and estimated force data points were collected. As shown in Fig.~\ref{fig:force_estimating}, the AFRG exhibits good estimation performance with a Root Mean Square Error (RMSE) of 0.23 N. The distributions of the forces are closely aligned, and the residuals follow an approximately normal distribution. In the scatter plot, most points are densely clustered, with the few outliers reflecting the short transient delay of the estimation during the force step changes at the beginning and end of each trial.

\subsection{Force Control Performance}
After validating the force estimation capability, the AFRG is employed to control the grinding process. Experiments are conducted in a factory environment to simulate realistic production conditions. The grinding trajectory follows a straight-line path along the workpiece edge, with target forces set to 2\,N, 3\,N, and 4\,N. Comparing the original and ground edges of the workpiece allows for a clear observation of grinding depth variations throughout the operation.

\begin{figure}[htbp] 
    \centering
    \includegraphics[width=0.95\linewidth]{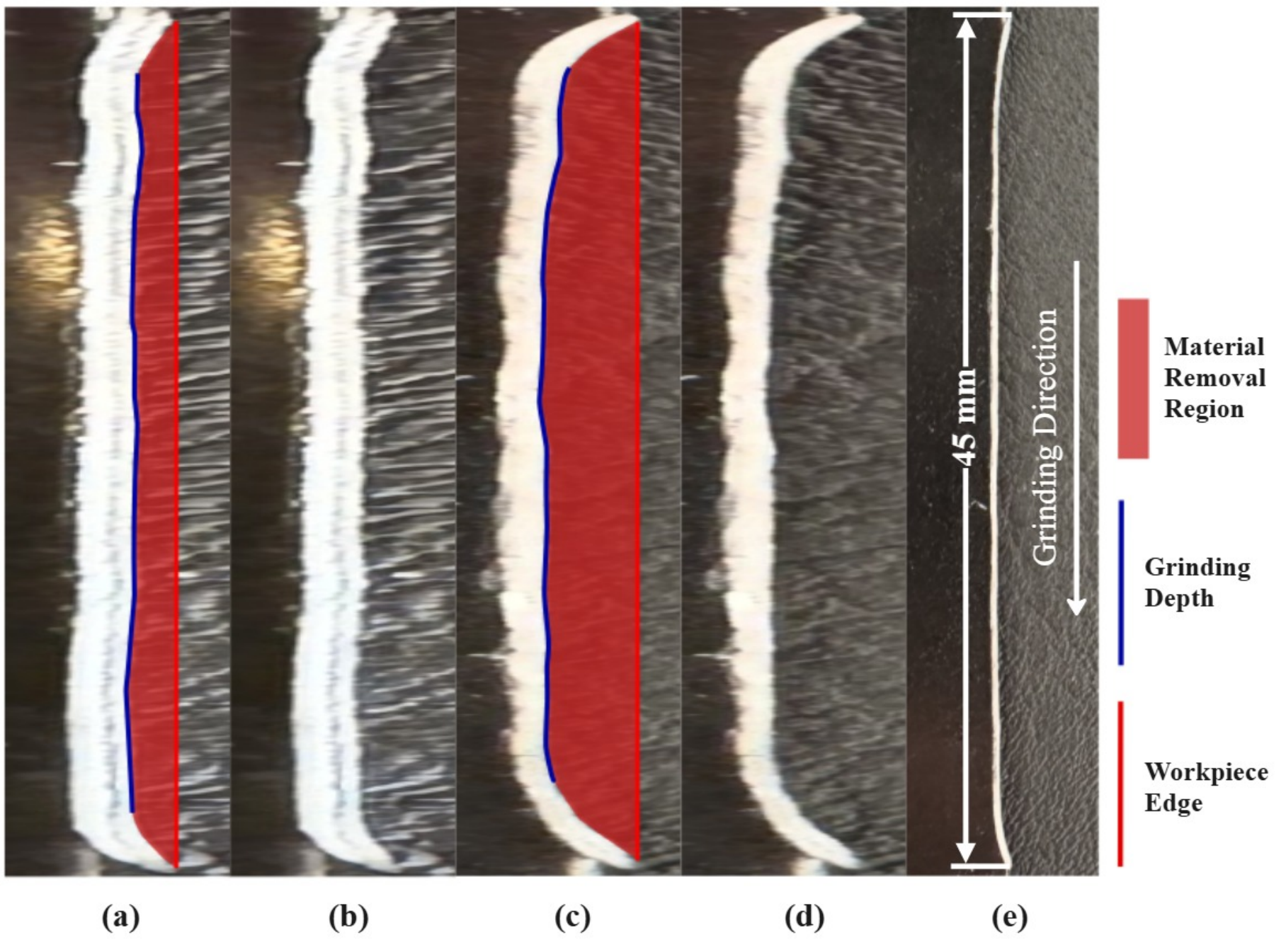}
    \caption{Workpieces (left side of each image) after grinding along a straight-line trajectory from top to bottom, horizontally stretched for clarity.\\
    (a, b) 3\,N target force: (a) annotated, (b) unannotated.\\
    (c, d) 4\,N target force: (c) annotated, (d) unannotated.\\
    (e) Image in original aspect ratio corresponding to (d).}
    \label{fig:sanding_path}
\end{figure}

Two post-grinding workpieces are shown in Fig.~\ref{fig:sanding_path}, where the target normal forces were set to 3\,N and 4\,N, respectively. From top to bottom, grinding was performed along a fixed feed rate $v_f$. In the figure, the red regions indicate the material removed from the workpiece, the red lines represent the original edges before grinding, and the blue lines correspond to the edges after grinding. Taking the red line as a reference, the blue line represents the grinding depth, denoted as $d$. Given the constant workpiece thickness $h$, the MRR can be expressed as
\begin{equation}
\dot{V} = d \cdot h \cdot v_f,
\end{equation}
which implies that the stability of grinding depth directly reflects the stability of MRR.  

To better visualise the grinding depth variation, Fig.~\ref{fig:sanding_path}(a–d) are horizontally stretched, while Fig.~\ref{fig:sanding_path}(e) shows the original scale. The near-linear blue lines indicate stable MRR during the grinding process.

\begin{figure}[htbp] 
    \centering
    \includegraphics[width=0.8\linewidth]{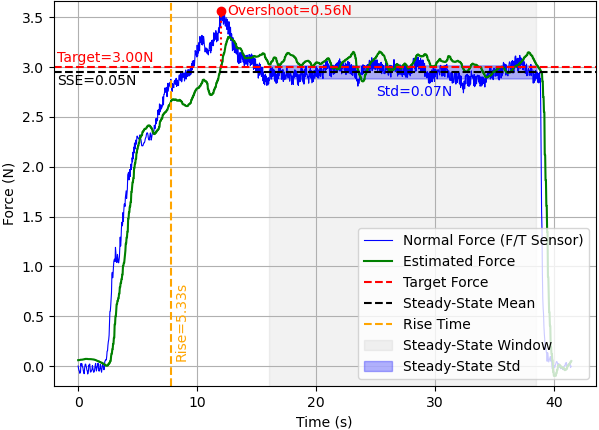}
    \caption{Force control performance of AFRG during a straight-line grinding. The measured normal force is compared to the target force, with annotations indicating rise time, overshoot, and steady-state error.}
    \label{fig:controller_performance}
\end{figure}

The force control performance is evaluated using the measurements from the F/T sensor. As shown in Fig.~\ref{fig:controller_performance}, the grinding force is maintained near the target force of 3\,N, with a Steady-State Error (SSE) of 0.05\,N and a standard deviation of 0.07\,N, indicating good control performance.

\begin{figure}[htbp] 
    \centering
    \includegraphics[width=0.8\linewidth]{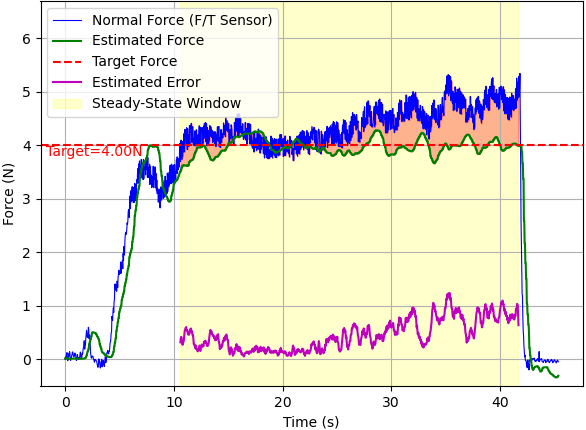}
    \caption{Absolute error between estimated and measured forces during straight-line grinding with AFRG, evaluated within the steady-state window.}
    \label{fig:controller_performance_2}
\end{figure}

The measured and estimated forces for the experiment corresponding to the workpiece shown in Fig.~\ref{fig:sanding_path}(c–e) are presented in Fig.~\ref{fig:controller_performance_2}, where the grinding disc was found to be worn at the end of the trial. The absolute estimated error between $\hat{F_n}$ and $F_n$ increases in the steady-state window, while the grinding depth remains stable. This occurs because the worn disc reduces $\mu_\theta$, causing $F_n$ to rise even though $\hat{F_n}$ remains near the target, as indicated by Eq.~(\ref{eq:x()}). 

The stability of the grinding depth indicates that the MRR remains stable, which can be explained by the fact that the relationship between $\omega$ and $F_t$,
\begin{equation}
F_t = \frac{\tau}{r} = \frac{f_m(\omega)}{r},
\end{equation}
is not affected by changes in $\mu_\theta$. Moreover, Li et al.~\cite{Li2024-tl} reported that $F_t$ is proportional to MRR and is not influenced by the grinding disc condition. Therefore,
\begin{equation}
\dot{V} \propto \hat{F_t} = \frac{f_m(\hat{\omega})}{r} = \mu_dx(s),
\label{eq:mrr_ft}
\end{equation}
where $\hat{F_t}$ is the estimated tangential force; $\mu_d$ is the approximate value of $\mu_\theta$ during data collection. Since $\mu_d$ was maintained as a constant during data collection, it can be treated as an invariant gain in this context. Therefore, variations in $\mu_\theta$ do not affect the relationship in Eq.~(\ref{eq:mrr_ft}). On the other hand, according to Eq.~(\ref{eq:mrr_fn}), changes in $\mu_\theta$ directly affect the relationship between $F_n$ and MRR, i.e., using F/T sensor-measured $F_n$ for control may result in MRR fluctuations due to variations in disc condition, whereas AFRG can maintain stable MRR.

\subsection{MRR Regulation Performance}
To evaluate the performance of the AFRG under disk-wear conditions, experiments were conducted on a stainless steel workpiece due to its high hardness. At a target force of 4\,N, the grinding disc experienced rapid wear, making this material suitable for assessing the AFRG's ability to maintain stable material removal under different disc conditions. Fig.~\ref{fig:tool_life}(b) illustrates the disc wear: a new disc is shown on the left, after a single 30\,s fixed-point trial in the middle, and after five trials (150\,s total) on the right.

\begin{figure}[htbp] 
    \centering
    \includegraphics[width=0.95\linewidth]{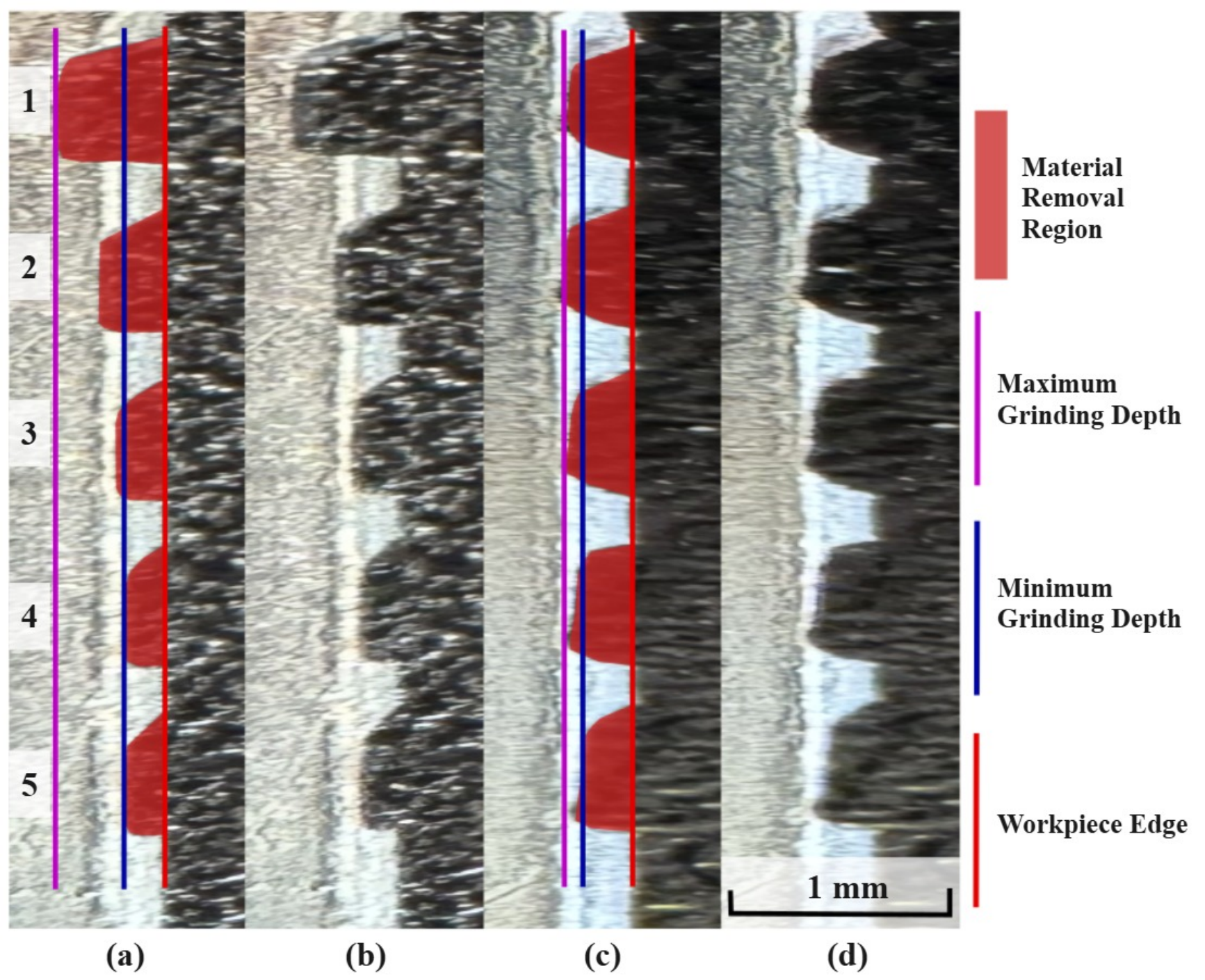}
    \caption{Workpiece surfaces after 1–5 fixed-point grinding trials, horizontally stretched for clarity.\\
    (a, b) F/T sensor-based force control: (a) annotated, (b) unannotated.\\
    (c, d) AFRG: (c) annotated, (d) unannotated.\\
    AFRG shows significantly improved MRR stability, with grinding depth variation approximately four times smaller compared to F/T sensor-based control.
    }
    \label{fig:sanding_depth}
\end{figure}

Two post-grinding workpieces are shown in Fig.~\ref{fig:sanding_depth}. Using the workpiece edge as a reference, the blue and purple lines indicate the maximum and minimum grinding depths, $d_{max}$ and $d_{min}$, respectively. In each experiment, a new grinding disc was used to conduct five fixed-point trials, each lasting $t_s$ seconds. The scratch width corresponds to the disc thickness, $w_d$. For a workpiece of uniform thickness $h$, the MRR can be expressed as
\begin{equation}
\dot{V} = \frac{w_d \, h \, d}{t_s}.
\end{equation}
Thus, the grinding depth serves as a reliable indicator of MRR.

Three repetitions were performed for each method: F/T sensor–based force control and AFRG. For the F/T method, the difference between $d_{max}$ and $d_{min}$ averaged 0.38\,mm across five trials. The first trial typically showed a higher MRR, which decreased as the trial progressed and the disc wore. In contrast, AFRG achieved an average difference of 0.09\,mm, one-fourth of that of the F/T method, demonstrating superior MRR stability.

\section{CONCLUSIONS}

This paper proposes AFRG, a data-driven robotic grinding method that relies solely on contact microphones as the sensing modality. Experiments demonstrated that AFRG can accurately estimate grinding force and successfully perform grinding tasks in a factory environment. Compared with conventional F/T sensor–based control, AFRG achieves a 4-fold increase in MRR consistency under varying disc conditions.

Furthermore, AFRG eliminates the need for force sensors, providing significant cost benefits. The contact microphone can be easily mounted on either the manipulator or the workpiece, enabling easy deployment. These characteristics make the approach particularly suitable for low-cost and rapid retrofitting of grinding robots that currently lack closed-loop force control, such as position-controlled robots.



However, AFRG has limitations. Grinding discs inevitably lose most of their cutting capability over time, leading to a decline in MRR. While disc replacement is required in such cases, our current system cannot autonomously detect this timing. Moreover, the current training process still requires an F/T sensor to provide ground-truth labels. Whether a model trained on one setup (e.g., a specific grinding tool or microphone placement) can generalise to other setups remains an open question. Future work will focus on integrating disc replacement timing prediction based on acoustic signal analysis and on improving model generalisation across different setups.


\section*{Acknowledgement}
This work was supported by the Australian Research Council under Grant No. IC200100001. The authors would like to acknowledge the support of Queensland University of Technology (QUT), the Australian Cobotics Centre (ACC), and the QUT Centre for Robotics (QCR). The authors further thank the Advanced Robotics for Manufacturing Hub (ARM Hub) for providing access to the experimental facilities.



\bibliographystyle{IEEEtran}  
\bibliography{reference}         

\end{document}